\documentclass[letterpaper, 10 pt, conference]{ieeeconf}

\usepackage[utf8]{inputenc}
\usepackage[T1]{fontenc}
\usepackage{acronym}

\usepackage{array}
\usepackage{float}
\usepackage{amsfonts}
\usepackage{amsmath}
\usepackage{amssymb}
\usepackage{mathtools}

\DeclareMathOperator*{\argmin}{\arg\!\min}
\DeclareMathOperator*{\argmax}{\arg\!\max}
\DeclarePairedDelimiterX{\infdivx}[2]{(}{)}{%
	#1\;\delimsize\|\;#2%
}
\newcommand{\infdiv}{D\infdivx}

\usepackage[english]{babel}

\usepackage{wrapfig}
\usepackage[export]{adjustbox}
\usepackage{microtype}
\usepackage[dvipsnames]{xcolor}
\usepackage{graphicx}
\usepackage{graphics}
\usepackage{svg}
\usepackage{tikz}
\usepackage[font=footnotesize]{caption}
\usepackage{subcaption}

\usepackage{multicol}
\usepackage{multirow}
\usepackage[font=footnotesize]{caption}
\usepackage{subcaption}
\usepackage{booktabs}

\usepackage{authblk}
\usepackage{algorithm, algorithmicx}
\usepackage[noend]{algpseudocode}

\usepackage{todonotes}
\usepackage{listings}

\renewcommand{\thetable}{\arabic{table}}

\graphicspath{{/images}}
\makeatletter
\def\endthebibliography{%
	\def\@noitemerr{\@latex@warning{Empty `thebibliography' environment}}%
	\endlist}
\makeatother

\definecolor{label-running} {RGB}{ 31,119,180}
\definecolor{label-walking} {RGB}{255,127, 14}
\definecolor{label-jumping} {RGB}{ 44,160, 44}
\definecolor{label-standing}{RGB}{148,103,189}
\definecolor{label-sitting} {RGB}{140, 86, 75}
\definecolor{label-lying}   {RGB}{127,127,127}
\definecolor{label-falling} {RGB}{188,189, 34}
\definecolor{label-transit} {RGB}{ 23,190,207}

\IEEEoverridecommandlockouts

\title{\LARGE \bf Multi-Task Reinforcement Learning in Continuous Control with Successor Feature-Based Concurrent Composition}
\author{Yu Tang Liu$^{1,2}$ and  Aamir Ahmad$^{2,1}$	\thanks{$^1$Max Planck Institute for Intelligent Systems, 72076 T{\"u}bingen, Germany. $^2$University of Stuttgart, 70569 Stuttgart, Germany. \texttt{yutang.liu@tuebingen.mpg.de, 
			aamir.ahmad@ifr.uni-stuttgart.de}}}

\makeatletter
\let\NAT@parse\undefined
\makeatother

\usepackage[english]{babel}
\usepackage{hyperref}

\begin{document}

	\maketitle 

	\makeatletter

\makeatother
\begin{abstract}
Deep reinforcement learning (DRL) frameworks are increasingly used to
solve high-dimensional continuous-control tasks in robotics. However, due to the lack of sample efficiency, applying DRL for online learning is still practically infeasible in the robotics domain. One reason is that DRL agents do not leverage the solution of previous tasks for new tasks. Recent work on multi-task DRL agents based on successor features (SFs) has proven to be quite promising in increasing sample efficiency. In this work, we present a new approach that unifies two prior multi-task RL frameworks, SF-GPI and value composition, and adapts them to the continuous control domain. 
We exploit compositional properties of successor features to compose a policy distribution from a set of primitives without training any new policy. Lastly, to demonstrate the multi-tasking mechanism, we present our 
proof-of-concept benchmark environments, Pointmass and Pointer, based on IsaacGym, which facilitates large-scale parallelization to accelerate the experiments. Our experimental results show that our multi-task agent has single-task performance on par with soft actor-critic (SAC), and the agent can successfully transfer to new unseen tasks. We provide our code as open-source at 
\url{https://github.com/robot-perception-group/concurrent_composition} for the 
benefit of the community. 
\end{abstract}



\section{Introduction}
\label{sec:Introduction}

One approach to achieving optimal continuous control in robotics is through 
reinforcement learning-based methods (RL) \cite{kober2013reinforcement}. 
However, if the reward definition changes, RL agents are usually optimized for one specific task and require more 
training. On the other hand, multi-task RL 
frameworks aim to design an RL agent that allows recycling old policies for 
other tasks to achieve higher sample efficiency \cite{lazaric2012transfer}. 

One way to achieve multi-task RL is through transfer learning by manipulating 
the policies trained on the old tasks. This work focuses on the transfer 
learning methods that define tasks using linearly decomposable reward 
functions, commonly used to define many robot control tasks \cite{ng1999policy}.

\begin{figure}[t!]
	\centering
	\includegraphics[width=0.4\textwidth]{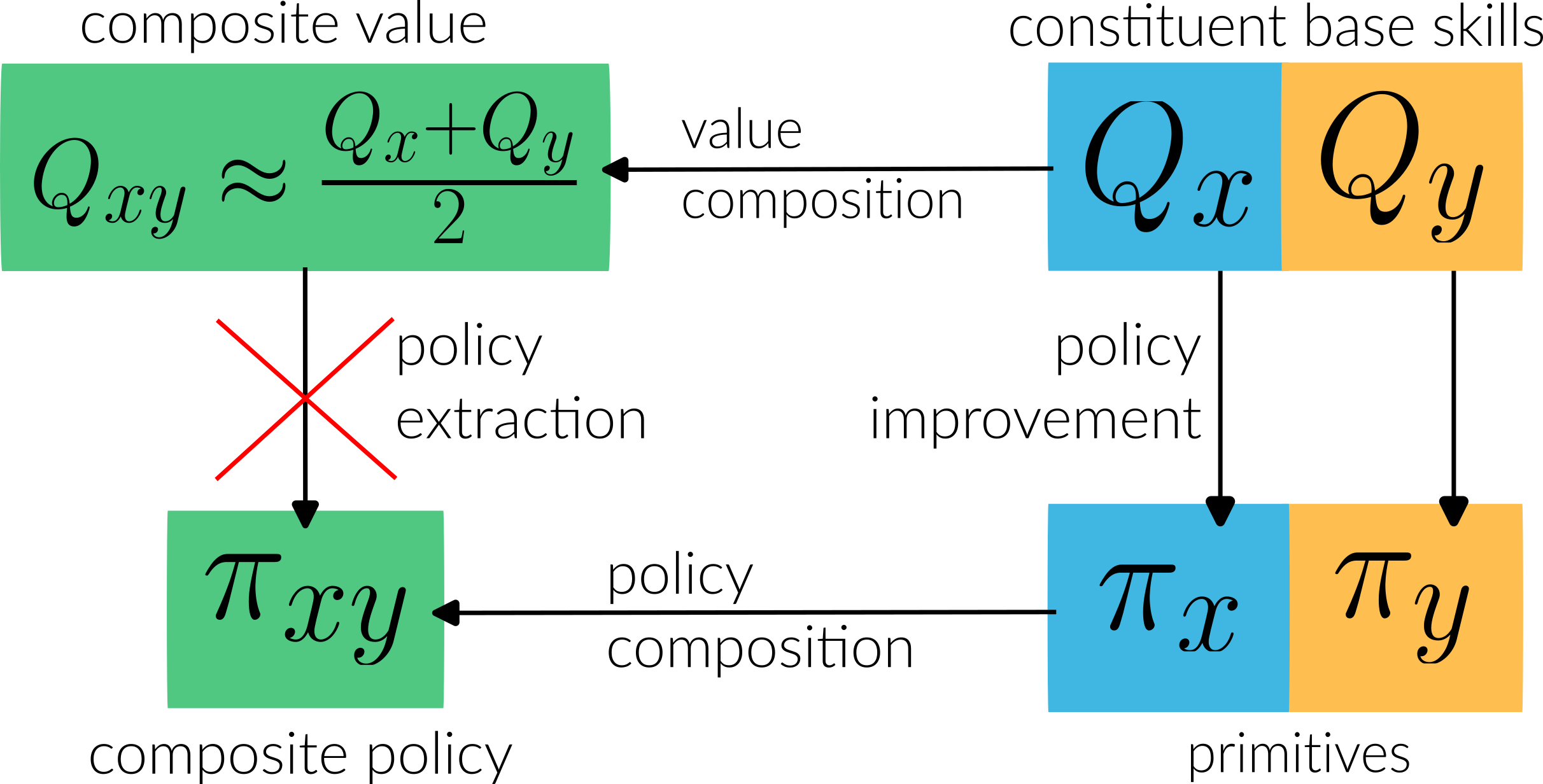}	
	\caption{In concurrent composition, policy extraction is intractable online. Instead, we propose composing the primitives directly in the run time.}
	\label{fig:illustration}
\end{figure}

Currently, there are two major transfer learning frameworks following the 
aforementioned assumption. The first framework, using \textit{successor 	features} and \textit{generalized policy improvement} (SF-GPI, \cite{barreto2020fast, barreto2019option}), derives the new policy by choosing 
the sub-policies, or primitives, with the largest associated action-value. The 
other framework directly composes the action-value of different tasks, which we 
refer to as \textit{value composition} (VC \cite{haarnoja2018composable, 	hunt2019composing, van2019composing, nangue2020boolean}). This framework first 
defines the composite value functions of new tasks by linearly combining the 
constituent action-value functions and then extracts the policy. 

Different from the \textit{Option} framework \cite{sutton1999between}, which is another approach to multi-task RL that combines the primitives in temporal order, in our novel approach, we focus on concurrent policy composition by composing the primitives at every time step. 
The main objective is to train primitives and enable the online composition to transfer knowledge for solving arbitrary tasks, thereby enhancing sample efficiency \cite{barreto2020fast}. Policy extraction is prohibitive since it requires updating policies multiple times. We consider real-time performance critical. 
Therefore, we must adapt VC-based methods for online composition. We do so by multiplicative policy composition (MCP \cite{peng2019mcp}). We show this can be achieved via the successor features framework, where SF-GPI becomes one special case.  

In summary, our contributions include the following:
\begin{itemize}
	\item A novel successor feature-based online concurrent composition framework unifying the GPI and VC methods.
	\item Deriving the relation between composition in value space and policy space.
	\item Extending this framework to composition in action space via the impact matrix.
	\item Introducing two benchmark multi-task environments, Pointmass and Pointer, based on IsaacGym \cite{liang2018gpu}.
\end{itemize}
 

\section{Related work}
\label{sec:2_related}

The compositionality in continuous control was first exploited by linearly-solvable Markov Decision Processes (LMDPs) \cite{todorov2006linearly}. It can be shown that composing the optimal value function of LMDPs results in the optimal value function in the composite task \cite{todorov2009compositionality}. However, LMDPs were restricted to known dynamics. 

The compositional agents in our framework instead build upon the two transfer learning frameworks: VC and SF-GPI. 
The \textbf{VC}-based methods often exploit the property that the optimal value function of conjunction tasks can be well approximated by the average of the constituent value functions \cite{haarnoja2018composable, hunt2019composing}. Interestingly, boolean 
operators can further expand the valid compositions in the discrete setting and 
achieve a super-exponential amount of task transfer \cite{van2019composing,nangue2020boolean}. However, none of these 
approaches consider concurrent composition scenarios. 

On the other hand, \textbf{SF-GPI} \cite{barreto2020fast} applies successor 
features \cite{dayan1993improving} to estimate a set of Q-functions in the new task. Maximizing these Q-functions results in policy improvement. The results in \cite{barreto2020fast} indicate that simultaneously training several primitives can significantly improve the sample efficiency for solving a new task with collective knowledge. Our work can be seen as an extension to SF-GPI by bringing valid operations from VC to successor feature-enabled concurrent learning.

%

\setlength{\belowdisplayskip}{0pt} \setlength{\belowdisplayshortskip}{0pt}
\setlength{\abovedisplayskip}{0pt} \setlength{\abovedisplayshortskip}{0pt}

\section{Background}
\label{sec:Background}

\subsection{Multi-Task Reinforcement Learning}
RL improves a policy through interaction with the environment, and it formulates this procedure by \textit{Markov decision processes} (MDPs), $\mathcal{M}\equiv<\mathcal{S},\mathcal{A},p,r,\gamma>$. $\mathcal{S}$ and $\mathcal{A}$ define the state space and action space respectively. The dynamics $p(s_{t+1}|s_t,a_t)$ defines the transition probability from one state to another. Reward function $r(s_t, a_t, s_{t+1})$ governs the desired agent behavior. The discount $\gamma \in [0,1)$ determines the valid time horizon of the task. The goal is to find the optimal control policy $\pi(a|s): S\rightarrow A$, such that the expected discounted return, $J(\pi)=\mathbb{E}_{\pi, \mathcal{M}}[\sum_{\tau=t}^{\infty} \gamma^{\tau-t}r_\tau]$, can be maximized. 

We consider transfer within a set of MDPs $\mathcal{M}_w$ where  $\mathcal{M}_w \equiv<\mathcal{S},\mathcal{A},p,r_w,\gamma>$, and each has its unique reward function, $r_w(s_t, a_t, s_{t+1})=\phi(s_t, a_t, s_{t+1})^\top \cdot w$, which is a linear combination of a set of commonly shared $d$ features, $\phi(s_t, a_t,s_{t+1})\in \mathbb{R}^d$ and the task weight $w\in \mathbb{R}^d$. The goal is to find a set of primitives that can solve the subset of tasks $w' \in \mathcal{W}'$ such that we can recover the policies of the unseen tasks $w \in \mathcal{W}$ by the found primitives, where $\mathcal{W}' \subset \mathcal{W}$.  

\subsection{Maximum Entropy Reinforcement Learning}
The maximum entropy RL framework \cite{ziebart2010modeling} introduces a  policy entropy bonus in the reward function. The goal is to find an optimal control policy that not only maximizes the trajectory reward but also encourages policy randomness, i.e.,
\begin{equation}
	\pi^* = \argmax_{\pi}\mathbb{E}_{\pi, \mathcal{M}_w}\left[ \sum_{\tau=t}^{\infty} \gamma^{\tau-t}(r_{w,\tau}+\alpha \mathcal{H}(\pi(\cdot|s_{\tau})))\right], 
\end{equation}
where $\alpha$ is a temperature scaler controlling the entropy bonus and $\mathcal{H}(\cdot)$ is an entropy estimator. Prior works have found that the entropy bonus can encourage exploration \cite{haarnoja2018soft}, increase the learning stability in noisy environments \cite{fox2015taming}, and preserve the policy multi-modality \cite{haarnoja2017reinforcement}. 

We define the surrogate objective: action-value function, or Q-function, of policy $\pi$ on task $w$ as:
\begin{equation}
	Q^{\pi}_{w}(s_t,a_t)\equiv r_{w,t}+\mathbb{E}_{\pi}\left[\sum_{\tau=t+1}^{\infty} \gamma^{\tau-t}(r_{w,\tau}+\alpha \mathcal{H}(\pi(\cdot|s_{\tau})))\right],
	\label{eqn:q function}
\end{equation}
and optimizing Q-function will result in an optimal maximum entropy policy. Note that when $\alpha=0$, we recover the standard RL Q-function. The Q-function can represent the performance of its correspondent policy, which can be improved recursively through the \textit{soft policy iteration}.

\subsection{Soft Policy Iteration}
The soft policy iteration can be defined in two steps: (1) \textit{soft policy evaluation} (2) \textit{soft policy improvement} \cite{haarnoja2017reinforcement}. The evaluation step estimates Q-function of the training policy through temporal difference learning, and it is formulated as follows, 
\begin{align}
	Q^{\pi}_{w}(s_t,a_t) & = r_w(s_t,a_t) + \gamma \mathbb{E}_{s_{t+1}\sim p}\left[ V^{\pi}_w(s_{t+1})\right] \label{Eqn:backup} \\
	V^{\pi}_{w}(s_t) &= \alpha \log \int_{a\in\mathcal{A}}e^{Q_w^{\pi}(s_t,a)/\alpha}. \label{Eqn:value backup}
\end{align}
Then the desired Boltzmann policy can be updated according to the soft policy improvement step,
\begin{equation}
	\pi_{desired}(a_t|s_t) = \frac{e^{Q_w^{\pi}(s_t,a_t)}}{\int_{a\in\mathcal{A}}e^{Q^{\pi}_{w}(s_t,a)}}  =
	\frac{e^{Q_w^{\pi}(s_t,a_t)}}{Z^{\pi}(s_t)}
	\propto e^{Q^{\pi}_w(s_t,a_t)},
\end{equation}
by solving the equation w.r.t each state,
\begin{equation}
	\pi(\cdot|s_t) =\argmin_{\pi'}\infdiv*{\pi'(\cdot|s_t)}{e^{Q^{\pi}_{w}(s_t,\cdot)}}, \label{eqn:soft policy improvement}
\end{equation}
where $D$ denotes the Kullback-Leiber divergence. Intuitively, the goal is to match the policy distribution to the Q-value of different actions; therefore, the higher the Q-value, the higher the probability of selecting the action. In practice, the integration above is replaced with the summation for the tracktability. And we approximate the Bolzmann policy by other distributions, e.g. a Gaussian $\pi = \mathcal{N}^{|\mathcal{A}|}$.

It can be shown that the evaluation step is a contraction with optimal Q-function as its fixed point, while the policy improvement step results in an improvement policy. Therefore, soft policy iteration will eventually converge to the optimal Q-function $Q^*_w$ on task $w$ with its corresponding optimal policy $\pi^*$, where $Q^*_w \equiv Q^{\pi^*}_w$ \cite{haarnoja2017reinforcement}.

\subsection{Successor Feature (SF)}
Given that  $r_w(s_t, a_t, s_{t+1})=\boldsymbol{\phi}(s_t, a_t, s_{t+1})^\top \cdot w$, we can rewrite the standard Q-function in following form, 
\begin{align}
	Q^{\pi}_w(&s_t,a_t)=\mathbb{E}_{\pi}\left[\sum_{\tau=t}^{\infty} \gamma^{\tau-t}r_{w,\tau}\right] \\
	&=\mathbb{E}_{\pi}\left[\sum_{\tau=t}^{\infty} \gamma^{\tau-t}\boldsymbol{\phi}^\top_t \right]\cdot w 
	=\boldsymbol{\psi}^{\pi}(s_t,a_t)^\top \cdot w \label{eqn:sf.w}
\end{align}
where $\boldsymbol{\psi}^{\pi}\in \mathbb{R}^d$ is the \textit{successor 
	feature} of policy $\pi$\cite{dayan1993improving}. 
Similar to Q-functions, which approximate "reward-to-go," SFs approximate the "feature-to-go" for a policy. Since Q-function can be viewed as a performance metric of a policy,  SFs provide a shortcut to estimate policy performance in different tasks $w$. In analogous to (\ref{Eqn:backup}, \ref{Eqn:value backup}), we can approximate SFs via soft policy iteration \cite{hunt2019composing}, i.e.
\begin{align}
	\boldsymbol{\psi}^{\pi}(s_t,a_t) 
	&= \boldsymbol{\phi}_t(s_t,a_t) + \gamma \mathbb{E}_{s_{t+1}\sim p} \left[ \boldsymbol{\Upsilon}(s_{t+1}) \right], \label{eqn:sf_evaluation}\\
	\boldsymbol{\Upsilon}^{\pi}(s_t) &= \alpha \log \int_{\mathcal{A}}e^{\boldsymbol{\psi}^{\pi}(s_t,a_t)/\alpha}. \label{eqn:sfv_evaluation}
\end{align}
The policy improvement step is similar to ($\ref{eqn:soft policy improvement}$) by replacing the Q-function with the approximation from the successor features and the desired primitive tasks, which is defined in Sec.~\ref{sec:Construct Policy Primitives}.

\subsection{Concurrent Composition}
During the training process for a specific task $w$, a single task RL agent always possesses a \textit{true} Q-function denoted as $Q^{True}_w$. However, $Q^{True}_w$ does not exist in a multi-task setting. Instead, it needs to be approximated using a set of primitive Q-functions $Q^{\Pi}_w=\{Q^{\pi_1}_w, ..., Q^{\pi_{n}}_w\}$ and value composition rules, i.e., $\hat{Q}^{Rule}_w=Rule(Q^{\Pi}_w)$, resulting in a composition loss, $|Q^{True}_w - \hat{Q}^{Rule}_w|$. One of our objectives is to identify a rule that minimizes this loss.

Similarly, in continuous control, a single task agent also possesses a \textit{true} policy denoted as $\pi^{True}_w$. The objective of policy composition is to approximate the true task policy $\pi^{True}_w$ by the composite $\hat{\pi}_w^{Rule}= Rule(\Pi,w)$ using a set of primitives $\Pi=\{\pi_1, ..., \pi_{n}\}$ and composition rules. We will show that for certain compositions in value space, corresponding rules exist in policy space. Furthermore, the composition loss in value space can lead to a performance loss in the composite policy, i.e., $Q^{\hat{\pi}^{Rule}}_w \leq Q^{\pi^{True}}_w$. This insight is crucial: the composite policy can only achieve performance equal to or lower than the single-task policy due to the incurred composition loss.

\subsection{Generalized Policy Improvement Composition (GPI)}
\label{sec: Generalized Policy Improvement Composition (GPI)}
When provided with primitives $\Pi$ and their corresponding Q-functions in a task $w$, denoted as $Q^{\Pi}_w$, a composite value, and policy can be constructed as follows:
\begin{align}
	Q^{True}_w(s,a) &\approx \hat{Q}^{GPI}_w(s,a) =  \max_{\pi \in \Pi} Q^{\pi}_{w}(s,a),\\
	\hat{\pi}^{GPI}(s) &\in \argmax_{a \in \mathcal{A}} \hat{Q}^{GPI}_w(s,a).
\end{align}

According to the GPI theorem \cite{barreto2020fast}, this new policy $\hat{\pi}^{GPI}$ is guaranteed to be at least as good as the primitives, as its Q-function is always greater than or equal to that of its members: $Q^{\hat{\pi}^{GPI}}_w(s,a) \geq \sup_{\pi \in \Pi} Q^{\pi}_w(s,a)$.

Successor features offer a quick and convenient method for evaluating the performance of a policy $\pi$ in any task $w$, i.e., $Q^{\pi_i}_w=w\cdot\psi^{\pi_i}$. This evaluation step is known as \textit{generalized policy evaluation} (GPE), and GPI can subsequently be applied to select the action that yields the highest value.

Note that our SF-GPI (Sec.~\ref{sec:GPI Composition}) sample actions from the primitives, i.e. $\argmax_{a \sim \Pi}$, since the $\argmax_{a\in \mathcal{A}}$ expression is infeasible in continuous action spaces.

\subsection{Value Composition (VC)}
\label{sec:Value Composition (VC)}
Combining the Q-functions provides a means to obtain the Q-function for a new task without explicitly solving for it. Previous studies have established a set of valid operations for multi-goal discrete control \cite{van2019composing, nangue2020boolean}. In continuous control, our focus is linear task combinations that do not assume binary task weights. Specifically, if a new task is a linear combination of old tasks, denoted as $r_{new} = b\cdot r_0+(1-b)\cdot r_1$, the optimal constituent Q-functions can approximate the optimal Q-function of the new task \cite{haarnoja2018composable}: $Q^{*}_{new} \approx \hat{Q}^{\sum}_{new}= b\cdot Q^*_0 + (1-b)\cdot Q^*_1$. The new policy can then be extracted from the composite. However, in continuous control, extracting process requires several policy improvement steps, making it intractable online. To extend this framework to concurrent composition, we relax the assumption and generalize the theorem to non-optimal value functions: $Q^{True}_{new} \approx \hat{Q}^{\sum}_{new}= b\cdot Q^{\pi_0}_0 + (1-b)\cdot Q^{\pi_1}_1$. With \ref{eqn:sf.w}, we can formulate VC to SFs framework.


\subsection{Multiplicative Compositional Policy (MCP)}
MCP  \cite{peng2019mcp}  has shown to compose the policy primitive in the 
imitation learning scenario. Here, we introduce it to concurrent composition. Given a set of $k$ Gaussian primitives $\Pi \in 
\mathcal{N}^{|\mathcal{A}|\times k}$ and a gating vector $g \in [0,1]^k$ 
where $||g||_1=1$, the function $MCP(\Pi, g): 
\mathcal{N}^{|\mathcal{A}|\times k} \times \mathbb{R}^k\rightarrow 
\mathcal{N}^{|\mathcal{A}|\times 1}$ is defined as follows,
\begin{equation}
	\pi(a|s) =MCP(\Pi, g) \equiv \frac{1}{Z(s)}\prod_{i=1}^{k} \pi_i(a|s)^{g_i},
\end{equation}
where $Z(s)$ is a partition function for normalization. If all the primitives are Gaussian, then the composite is also a Gaussian. 
It was not explained why such composition is valid. However, we show this is a natural outcome from value composition in the policy space (Sec.~\ref{sec:MSF Composition}). 

\section{Methodology}
\label{sec:Methodology}
We present our novel approach in the following sections, which are eventually summarized in Algorithm~\ref{alg:summary}.


\subsection{Successor Feature-based Composition}
\label{sec:Successor Feature-based Composition}
We introduce two more SF-based compositions and our GPI composition rule (Sec.~\ref{sec:GPI Composition}).

\subsubsection{Successor Feature based Value Composition (SFV)}
First, we introduce the linear value composition to the successor feature framework. Given the task $r_w=\boldsymbol{\phi}^\top \cdot w$, we can treat features $\boldsymbol{\phi}$ as sub-tasks. From this perspective, $\boldsymbol{\phi}$ becomes the constituent sub-tasks, and $r_w$ is the composite task. Thus, the successor features represent the primitive's performance on each sub-task, e.g., $\psi^{\pi_i}_j$ is the performance metric of $\pi_i$ on controlling sub-task $\phi_j$. Therefore, (\ref{eqn:sf.w}) can be seen as a special case for the value composition of the same policy.

Value composition theorems state that the average of the Q-functions is a sufficiently good approximation of the true Q-function \cite{haarnoja2018composable,van2019composing, nangue2020boolean}. We can define the composite Q-function and discrete greedy policy accordingly,
\begin{align}
	Q^{True}_w(s,a) \approx \hat{Q}^{{SFV}}_w &\equiv  \frac{1}{n} \sum_{i=1}^{n} Q_w^{\pi_i}(s,a)= \frac{\boldsymbol{w}^{\top}}{n} \sum_{i=1}^{n} \boldsymbol{\psi}^{\pi_i}(s,a)
	\label{eqn:SFV value} \\
	\hat{\pi}^{SFV}(s) &\in \argmax_{a \in \mathcal{A}} 
	\hat{Q}^{{SFV}}_w(s,a).   \label{eqn:SFV policy}  
\end{align}
This expression shows that SFV composition aims to achieve an average performance in all sub-tasks. 

\subsubsection{Maximum Successor Feature Composition (MSF)}
We can treat successor features as sub-tasks. In that case, it naturally raises the question: can we combine primitives in a certain way to achieve the current best achievable performance in all sub-tasks and, subsequently, any composite task? That is, given SFs of their corresponding $n$ primitives, $\Psi^{\Pi}=\{\boldsymbol{\psi}^{\pi_1},..., \boldsymbol{\psi}^{\pi_{n}}\}$, where $\boldsymbol{\psi}^{\pi_i}=\{\psi_1^{\pi_i}, ...\psi_{d}^{\pi_i} \}$. Then the desired composition operator $\mathcal{U}$ can be defined as $\mathcal{U}\Psi^{\Pi}\equiv \{\max_{\pi \in \Pi}{\psi_1^{\pi}},...,\max_{\pi \in \Pi}{\psi_{d}^{\pi}} \}$. With this, we define the composition and its discrete greedy policy
\begin{align}
	Q^{True}_w(s,a) &\approx \hat{Q}^{{MSF}}_w(s,a) \equiv w^\top\mathcal{U} \Psi^{\Pi}(s,a)   \label{eqn:MSF composition}    \\
	\hat{\pi}^{MSF}(s) &\in \argmax_{a \in \mathcal{A}} \hat{Q}^{{MSF}}_w(s,a).   \label{eqn:MSF policy}    
\end{align}

To summarize, all the proposed composition rules either overestimate or underestimate the true Q-function. Specifically, they exhibit the following relationships:
\begin{equation}
	\hat{Q}^{MSF}_w \geq \hat{Q}^{SFV}_{w} \geq Q^{True}_{w}  \geq \hat{Q}^{GPI}_w .
\end{equation}
Like VC, both SFV and MSF assume that all sub-tasks can be accomplished simultaneously, leading to an overestimation when this assumption does not hold. Conversely, GPI assumes that only one task can be achieved at a time, resulting in an underestimation.
Due to the imperfect nature of composition rules, the composites are rarely optimal, i.e.,
\begin{equation}
	Q^{\pi^{True}}_w \geq \{Q^{\hat{\pi}^{SFV}}_{w}, \hat{Q}^{\hat{\pi}^{MSF}}_{w}, Q^{\hat{\pi}^{GPI}}_w\} .
\end{equation}

\subsection{Construct Primitives}
\label{sec:Construct Policy Primitives}
We train primitives based on the sub-tasks, and each primitive is only responsible for its sub-task. Therefore, we have the same number of sub-tasks and primitives (i.e., $d=n$). All primitives are trained by soft policy iterations and follow its convergence guarantee, i.e.
\begin{align}
	\pi_i(\cdot|s) \propto e^{Q^{\pi_i}_{\mathbf{e}_i}(s,\cdot)}=e^{\mathbf{e}_i^{\top}\cdot\psi^{\pi_i}(s,\cdot)}.  \label{eqn:base_tasks}
\end{align}
The base tasks $\mathbf{e}_i \in \mathbb{R}^d$ can be constructed arbitrarily, as long as the base task and the transfer task share the same sub-tasks or overlap in value space. For example, the base task defined as the zero vector will fail since this task has no overlapping with any transfer task. For simplicity, we define $\mathbf{e}_i$ by elementary vector such that the primitives form the "basis" in the policy space. 

\subsection{Approximate True Policy by Composing Primitives}
\label{sec:Approximate True Policy by Composing  Primitives}
This is the core part of our method. We provide an analytical view of how the true policy can be reconstructed by a set of primitives given their relation in value space introduced in Sec.~\ref{sec:Successor Feature-based Composition}.

\subsubsection{MSF Composition}
\label{sec:MSF Composition}
Following MSF in value space (\ref{eqn:MSF policy}), we derive the correspondent policy composition,
\begin{align}
	\pi^{True}(\cdot|s) &\propto e^{Q^{True}_w(s,\cdot)}
	\approx e^{\hat{Q}^{{MSF}}_w(s,a)} = e^{w^\top\mathcal{U} \Psi^{\Pi}(s,a)}
	\label{eqn:MSF approx}\\
	&\approx e^{w^{\top}\cdot [ \psi^{\pi_1}_1,...,\psi^{\pi_{d}}_{d} ] (s,\cdot)} 
	\propto \prod_{i=1}^{d} \pi_i(\cdot|s)^{w_i}. \label{eqn:MSF Composition}
\end{align}
The approximation in (\ref{eqn:MSF Composition}) assumes the maximum SF of each sub-task is most likely to be the SF of the primitive trained from the respective sub-task. Then the MSF composition can be achieved by MCP, i.e., $\hat{\pi}(\cdot|s) = MCP(\Pi, g)$, where $g = \frac{w}{|w|}$  is the normalized gating vector.

\subsubsection{SFV Composition}
Following SFV in value space (\ref{eqn:SFV value}), we derive the correspondent policy composition,
\begin{align}
	&\pi^{True}(\cdot|s) \propto e^{Q^{True}_w(s,\cdot)}\approx e^{\hat{Q}^{SFV}_w(s,\cdot)}  
	=  \prod_{i=1}^{d} e^{\frac{\boldsymbol{w}^{\top}}{d} \boldsymbol{\psi}^{\pi_i}(s,\cdot)} \\
	&=  \prod_{i=1}^{d} e^{\frac{1}{d}  w_i \psi_i^{\pi_i}(s,\cdot)}  \prod_{i=1}^{d}  \prod_{j\neq i}^{d} e^{\frac{1}{d}  w_j \psi_j^{\pi_i}(s,\cdot)}   
	\propto \prod_{i=1}^{d} \pi_i(\cdot|s)^{\frac{w_i}{d}}\cdot \epsilon  
\end{align}
where $\epsilon$ is the artefactual noise. Because of the noise, the SFV composition is valid but impractical in policy space with the set of primitives introduced in Sec.~\ref{sec:Construct Policy Primitives}. 

\subsubsection{GPI Composition}
\label{sec:GPI Composition}
Though this is already an SF-based composition, we reformulate it for completeness.
\begin{align}
	\pi^{True}(\cdot|s) \propto e^{Q^{True}_w(s,\cdot)} &  \approx e^{\hat{Q}^{GPI}_w(s,\cdot)} \\
	= e^{\max_{\pi_i \in \Pi}Q^{\pi_i}_w(s,\cdot)}  
	& \propto \argmax_{\pi_i \in \Pi}Q^{\pi_i}_w(s,\cdot)
\end{align}
The approximation can be achieved by successor features, i.e. $\hat{\pi}(\cdot|s)=\argmax_{\pi_i \in \Pi}Q^{\pi_i}_w(s,\cdot)=\argmax_{\pi_i \in \Pi}w^\top\psi^{\pi_i}(s,\cdot)$. 

We summarize the above policy composition rules in Algorithms~\ref{alg:msf}~\ref{alg:gpi}. For clarity, our proposed approach incorporates two transfer mechanisms that complement each other to expedite learning. First, SFs facilitate transfer to any task by specifying different task weights, enabling policy evaluation without additional training. This transfer is associated with composition in the value space. Second, as the composite policy outperforms the primitives, the trajectories gathered by the composite explore higher-value regions in the composite task. This transfer mechanism resembles \textit{teacher-student} \cite{schmitt2018kickstarting} or \textit{learn-from-demonstration} \cite{hester2018deep} approaches, where knowledge is transferred from the expert policy to the primitives.

At this point, we have constructed an analytic method to find the policy composition corresponding to the composition in value space. Note that if primitives are constructed with different base tasks $\mathbf{e}_i$, the composition in policy space will change accordingly. Next, we heuristically extend the idea to \textit{direct action composition}, which allows composition to happen in the action space.
\begin{algorithm}[h]
	\caption{MSF Composition \label{alg:msf}}
	\begin{algorithmic}[1]
		\Function{MSF}{s, w}:
		\State $\hat{\pi}(\cdot|s) = MCP(\Pi(\cdot|s), \frac{w}{|w|})$ 
		\State $a\sim \hat{\pi}(\cdot|s)$ \Comment{sample from composite}
		\State \Return $a$
		\EndFunction		
	\end{algorithmic}
\end{algorithm}
\begin{algorithm}[h]
	\caption{GPI Composition \label{alg:gpi}}
	\begin{algorithmic}[1]
		\Function{SF-GPI}{s, w}:
		\State $\boldsymbol{a}\sim \Pi(\cdot|s)$ \Comment{sample from primitives}
		\State $\hat{\boldsymbol{Q}}^{\pi}	_w = \boldsymbol{\psi}^{\pi}(s,\boldsymbol{a})\cdot w$ \Comment{GPE}
		\State $a = \argmax_{a \sim \Pi} \hat{\boldsymbol{Q}}^{\pi}_w$ \Comment{GPI}
		\State \Return $a$
		\EndFunction		
	\end{algorithmic}
\end{algorithm}

\subsection{Multiplicative Compositional Action (MCA)}
\label{sec:Multiplicative Compositional Action}
Rewriting a Gaussian primitive $\pi_i \in \Pi$ in the form,
\begin{align}
	\pi_i(a|s)=[\mathcal{N}^{\pi_i}_1, ...,\mathcal{N}^{\pi_i}_{|\mathcal{A}|}]
\end{align}
where $\mathcal{N}^{\pi_i}_j=\mathcal{N}^{\pi_i}(\mu_j,\sigma_j)$ is the shorthand for the j-th action component of $\pi_i$ . Then the MSF composition ($\ref{eqn:MSF Composition}$) has the action components in the following form,
\begin{align}
	\hat{\pi}(\cdot|s) = MCP(\Pi, g)=[\prod_{i=1}^{d} {(\mathcal{N}^{\pi_i}_1)^{g_i}}, ...,  \prod_{i=1}^{d} {(\mathcal{N}^{\pi_i}_{|\mathcal{A}|})^{g_i}} ].
\end{align}

However, since all action components are coupled by the exponent $g_i$, we cannot manipulate them differently. Alternatively, given $\kappa\in [0,1]^{|\mathcal{A}|\times n}$ where $||\kappa||_1=1$, we define MCA that applies composition in the action space $MCA(\Pi, \kappa): \mathcal{N}^{|\mathcal{A}|\times n} \times \mathbb{R}^{|\mathcal{A}|\times n}\rightarrow \mathcal{N}^{|\mathcal{A}|\times 1}$:
\begin{align}
	\hat{\pi}(\cdot|s) = MCA(\Pi, \kappa) \equiv [\prod_{i=1}^{d} {(\mathcal{N}^{\pi_i}_1)^{\kappa^{\pi_i}_1}}, ...,  \prod_{i=1}^{d} {(\mathcal{N}^{\pi_i}_{|\mathcal{A}|})^{\kappa^{\pi_i}_{|\mathcal{A}|}}} ]. \label{eqn:mca}
\end{align}

We define $\kappa^{\pi_i}_j$  in such a way that action components with lower successor feature value will be filtered out and higher value to be emphasized, i.e., $	\kappa^{\pi_i} \propto \mathcal{P}_{|\mathcal{A}|\times d} \cdot \boldsymbol{\psi}^{\pi_i}$, where $\mathcal{P}$ is an \textit{impact matrix} defining the mapping from successor feature to the action components. 

\subsection{Impact Matrix}
\label{sec:impact matrix}
We would like to know the relationship between each action component and the features throughout the trajectory, which is, surprisingly, the successor feature. One way to determine the relationship between two variables is by observing the change in one variable when the other is tuned, which can be achieved through derivatives. Given a set of successor features $\boldsymbol{\psi}=\{\psi_1, ..., \psi_{d}\}$, we define the impact matrix as the absolute value of the Jacobian of $\psi$ w.r.t the action components:
\begin{gather}
	\mathcal{P}\equiv |\nabla_a{\psi(s,a)}|=
	\begin{bmatrix}
		|\frac{\partial \psi_1}{\partial a_1}|  & ... & |\frac{\partial \psi_{d}}{\partial a_1} | \\
		\vdots & \ddots & \\
		|\frac{\partial \psi_1}{\partial a_{|\mathcal{A}|}}|  & ... & |\frac{\partial \psi_{d}}{\partial a_{|\mathcal{A}|}}|  
	\end{bmatrix}
	_{|\mathcal{A}| \times d}
\end{gather}

The computation required by this expression and the accuracy of the derivative are major concerns. Fortunately, in practice, this estimation can be efficiently computed using vectorized operations with GPUs. The accuracy of the derivative improves over time as the successor features become smoother during training. However, in the early stages of training, the estimation can be noisy. We can mitigate this noise by averaging over all policies, i.e., $\boldsymbol{\psi} \approx \frac{1}{n}\sum_n{\boldsymbol{\psi}^{\pi_n}}$. Consequently, the estimation becomes less noisy with an increase in the number of primitives. To further reduce noise, we incorporate temporal dependency by averaging with the impact matrix from the previous time step, i.e., $\hat{\mathcal{P}} \approx (\mathcal{P}_{t-1}+\mathcal{P}_{t})/2$.

\subsection{Direct Action Composition (DAC)}
\label{sec:Direct Action Composition}
With the impact matrix, we define $\kappa \equiv \mathcal{P}_{|\mathcal{A}|\times d} \cdot \Psi_{d \times n}\odot w_{d\times 1}$, where  
\begin{gather}
	\Psi=\{\boldsymbol{\psi}^{\pi_1},..., \boldsymbol{\psi}^{\pi_{n}}\}=
	\begin{bmatrix}
		\psi_1^{\pi_1}  & ... & \psi_1^{\pi_{n}}  \\
		\vdots & \ddots & \\
		\psi_{d}^{\pi_1}  & ... & \psi_{d}^{\pi_{n}} 
	\end{bmatrix}
	_{d\times n},
\end{gather}
$\odot$ denotes the Hadamard product, and $n=d$. Intuitively, task weights $w$ filter the task-relevant successor features and then map their impact to each action component via the impact matrix, i.e.,
$\hat{\kappa}=\mathcal{P}\cdot \Psi\odot w$. And then DAC can be achieved by $MCA$ (\ref{eqn:mca}) as shown in Algorithms~\ref{alg:dac}. We introduce another variant, DAC-GPI, by choosing the action component with the largest $\hat{\kappa}$, as displayed in Algorithms~\ref{alg:dacgpi}.
\begin{algorithm}
	\caption{DAC Composition} \label{alg:dac}
	\begin{algorithmic}[1]
		\Function{DAC}{s, w}:
		\State $\boldsymbol{a} \sim \Pi(\cdot|s)$ \Comment{sample from primitives}
		\State $\hat{\kappa}=\mathcal{P}\cdot \Psi(s,\boldsymbol{a}) \odot w$ 
		\State $\hat{\pi}(\cdot|s) = MCA(\Pi(\cdot|s), \frac{\hat{\kappa}}{|\hat{\kappa}|})$ 
		\State $a\sim \hat{\pi}(\cdot|s)$ \Comment{sample from composite}
		\State \Return $a$
		\EndFunction		
	\end{algorithmic}
\end{algorithm}
\begin{algorithm}
	\caption{DAC-GPI Composition} \label{alg:dacgpi}	
	\begin{algorithmic}[1]
		\Function{DAC-GPI}{s, w}:
		\State $\boldsymbol{a} \sim \Pi(\cdot|s)$ \Comment{sample from primitives}
		\State $\hat{\kappa}=\mathcal{P}\cdot \Psi(s,\boldsymbol{a}) \odot w $ 
		\State $a = [\argmax_{a_1} \hat{\kappa}_{1}^{\pi}, ..., \argmax_{a_{|\mathcal{A}|}} \hat{\kappa}_{|\mathcal{A}|}^{\pi} ]$ \Comment{GPI}
		\State \Return $a$
		\EndFunction		
	\end{algorithmic}
\end{algorithm}

In practice, we replace $\Psi$ with advantage $\Gamma$ for numerical stability, i.e. $\Gamma=\Psi - \frac{1}{n}\sum_n{\boldsymbol{\psi}^{\pi_n}}$. Empirically, we found the performance much better when we clip the advantage below zero, i.e., $\Gamma_{clip}=max(0,\Gamma)$. Finally, we have $\hat{\kappa}=\mathcal{P}\cdot \Gamma_{clip}\odot w$.

Notice that DAC-GPI reduces to SF-GPI when all the action components come from the same policy. Moreover, when each primitive shares the same exponent across all action components, DAC reduces to MSF. The major difference between DAC and DAC-GPI is that the former composes action components from all primitives according to the SFs, while the latter only selects the best. 

In summary, we have constructed a method to bridge the composition in value space to policy space. This provides an effective analytic method for generating different concurrent composition algorithms. We then show that the composition can happen in policy space and action space with the help of the impact matrix. Unlike standard RL agents, compositional agents usually involve critics or SFs in the action selection process and filter out the lower-quality actions before delivery.

\subsection{Training the Network}
The training data is gathered by direct interaction with the environment through composition rules. Therefore, we can treat the concurrent compositional agent just as any regular RL agent and train end-to-end. 

In total, we have $n$ pairs of SFs and primitives with the $i$-th successor feature network $\psi^{\pi_i}$, target network $\bar{\psi}^{\pi_i}$, and the primitive network $\pi_i$ paramterized by $\theta_{\psi^{\pi_i}}$, $\theta_{\bar{\psi}^{\pi_i}}$, and $\theta_{\pi_i}$ respectively. The training scheme follows the soft policy iteration, which includes the evaluation step (\ref{eqn:sf_evaluation},\ref{eqn:sfv_evaluation}) and improvement step (\ref{eqn:soft policy improvement}). The evaluation step now evaluates the successor feature instead of the Q-function. In contrast, the policy improvement step stays the same by following the base task Q-function $Q^{\pi_i}_{\mathbf{e}_i}$. Therefore, the loss for the i-th successor feature can be computed as follows,
\begin{align}
	J_{\boldsymbol{\psi}^{\pi_i}}(\theta_{\boldsymbol{\psi}^{\pi_i}}) &= \mathbb{E}_{(s_t,a_t)\sim\mathcal{D}}    \left[     ( \boldsymbol{\psi}^{\pi_i}(s_t,a_t) - \hat{\boldsymbol{\psi}}^{\pi_i}(s_t,a_t)  )^2	\right],
\end{align}
where $\mathcal{D}$ is the replay buffer and we have the TD-target $\hat{\boldsymbol{\psi}}^{\pi_i}$ approximated by the target network $\bar{\psi}^{\pi_i}$,
\begin{align}
		\hat{\boldsymbol{\psi}}^{\pi_i}(s_t,a_t)&=  \boldsymbol{\phi}_t + \gamma \mathbb{E}_{s_{t+1}\sim p(\cdot|s_t,a_t)} \left[ \boldsymbol{\Upsilon}^{\pi_i}(s_{t+1}) \right], \\
		\boldsymbol{\Upsilon}^{\pi_i}(s_t) &=   \mathbb{E}_{a_t \sim \pi_i}\left[  \boldsymbol{\bar{\psi}}^{\pi_i}(s_t,a_t)-\log \pi_i(a_t|s_t) \right].
\end{align}
where $- \alpha \log \pi$ is a one-step policy entropy estimator. Then the primitive can be improved by reducing the KL-divergence in (\ref{eqn:soft policy improvement}), which can be approximated by maximizing the objective,
\begin{align}
	J_{\pi_i}(\theta_{\pi_i})=\mathbb{E}_{s_t\sim \mathcal{D}, a_t \sim \pi_i} \left[ Q^{\pi_i}_{\mathbf{e_i}}(s_t, a_t) -\log \pi_i( a_t|s_t )   \right],
\end{align}
with the primitive Q-function computed by (\ref{eqn:base_tasks}). 

We summarize our approach in Algorithm~\ref{alg:summary}. The method alternates between the interaction phase for collecting samples by the compositional policy and the learning phase that updates the function approximators' parameters by stochastic gradient descend using the collected samples. Although looping over all primitives may appear intimidating in the algorithm, in our implementation, we vectorized computations to update all primitives simultaneously in a single pass such that the computation time remains constant.
\begin{algorithm}
	\caption{Composition Agent}\label{alg:summary}
	\begin{algorithmic}[1]
		\Statex Choose $method\in$ MSF, SF-GPI, DAC, DAC-GPI
		\Statex Initialize network parameters $\boldsymbol{\theta_{\psi^{\pi}}}, \boldsymbol{\theta_{\bar{\psi}^{\pi}}}, \boldsymbol{\theta_{\pi}}$

		\For{N steps} 
			\If{exploration} $a\leftarrow$ Uniform($\mathcal{A}$)  
			\Else $~a\leftarrow method(s)$ 				
			\EndIf
			\State $\mathcal{D}\leftarrow \mathcal{D} \cup (s_t,a_t,\phi(s_t,a_t), s_{t+1})$  

			\For{each gradient step}
				\For{$i \leftarrow 1,2,...,n$}
					\State $\theta_{\psi^{\pi_i}}\leftarrow \theta_{\psi^{\pi_i}}- \lambda_{\psi} \nabla_{\theta_{\psi^{\pi_i}}}J_{\psi^{\pi_i}}(\theta_{\psi^{\pi_i}}) $
					\State $\theta_{\pi_i}\leftarrow \theta_{\pi_i}- \lambda_{\pi} \nabla_{\theta_{\pi_i}}J_{\pi_i}(\theta_{\pi_i}) $
					\State $\theta_{\bar{\psi}^{\pi_i}} \leftarrow \tau \theta_{\bar{\psi}^{\pi_i}} + (1-\tau) \theta_{\psi^{\pi_i}}$ 
				\EndFor
			\EndFor
		\EndFor
	\end{algorithmic}
\end{algorithm}

 

\section{Experiments and Results}
\label{sec:4_experiment}
The experiment aims to test if the multi-task RL agents generated by the proposed framework, i.e., SF-GPI, MSF, DAC, DAC-GPI, can be task-agnostic and transfer to unseen tasks. 

\begin{figure}[b]
	\centering
	\begin{subfigure}[b]{0.35\textwidth}
		\centering
		\includegraphics[width=0.8\textwidth]{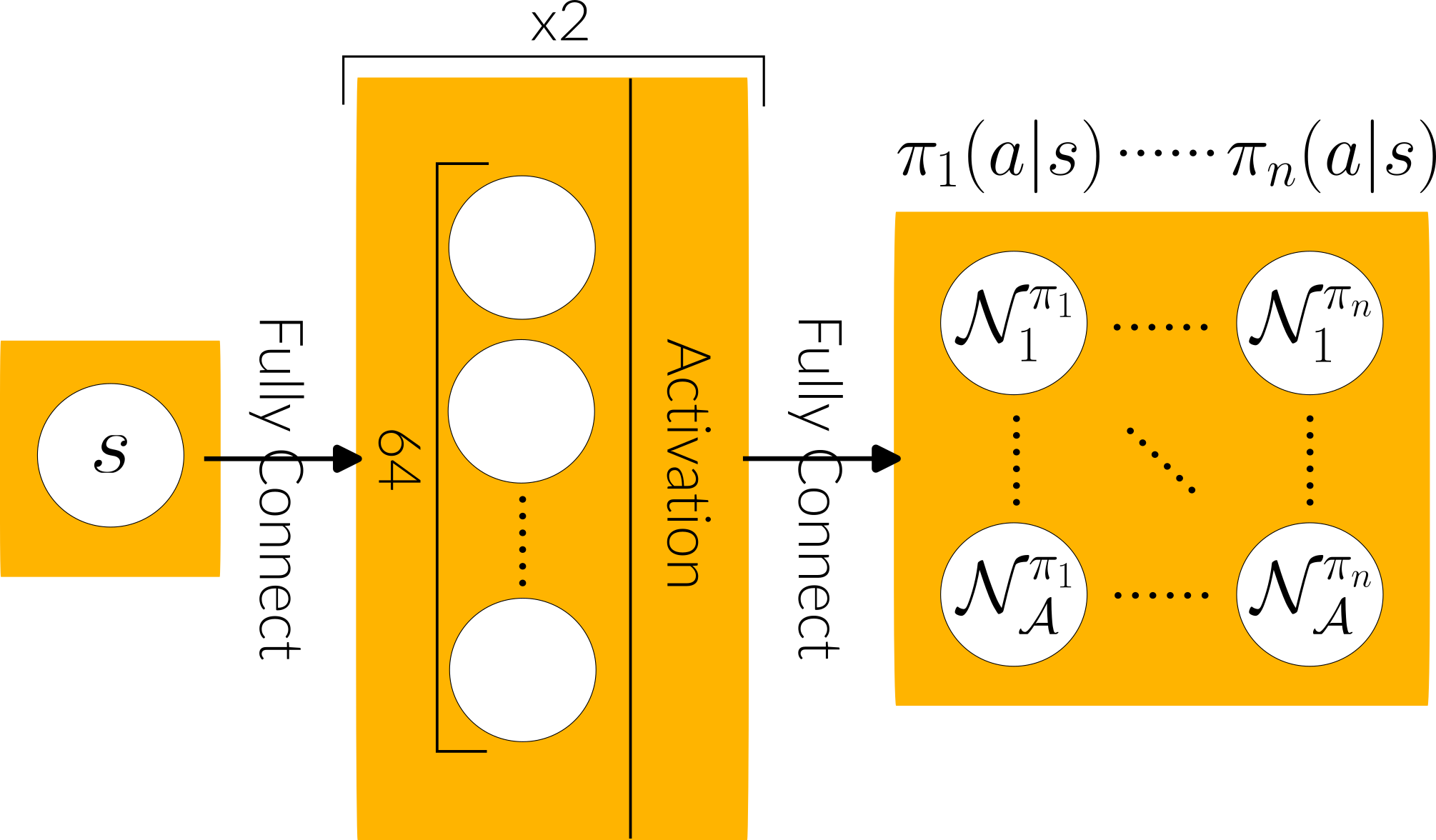}  
		\caption{primitive network.}
		\label{fig:pi_connect}
	\end{subfigure}
	\hfill
	\begin{subfigure}[b]{0.35\textwidth}
		\centering
		\includegraphics[width=0.95\textwidth]{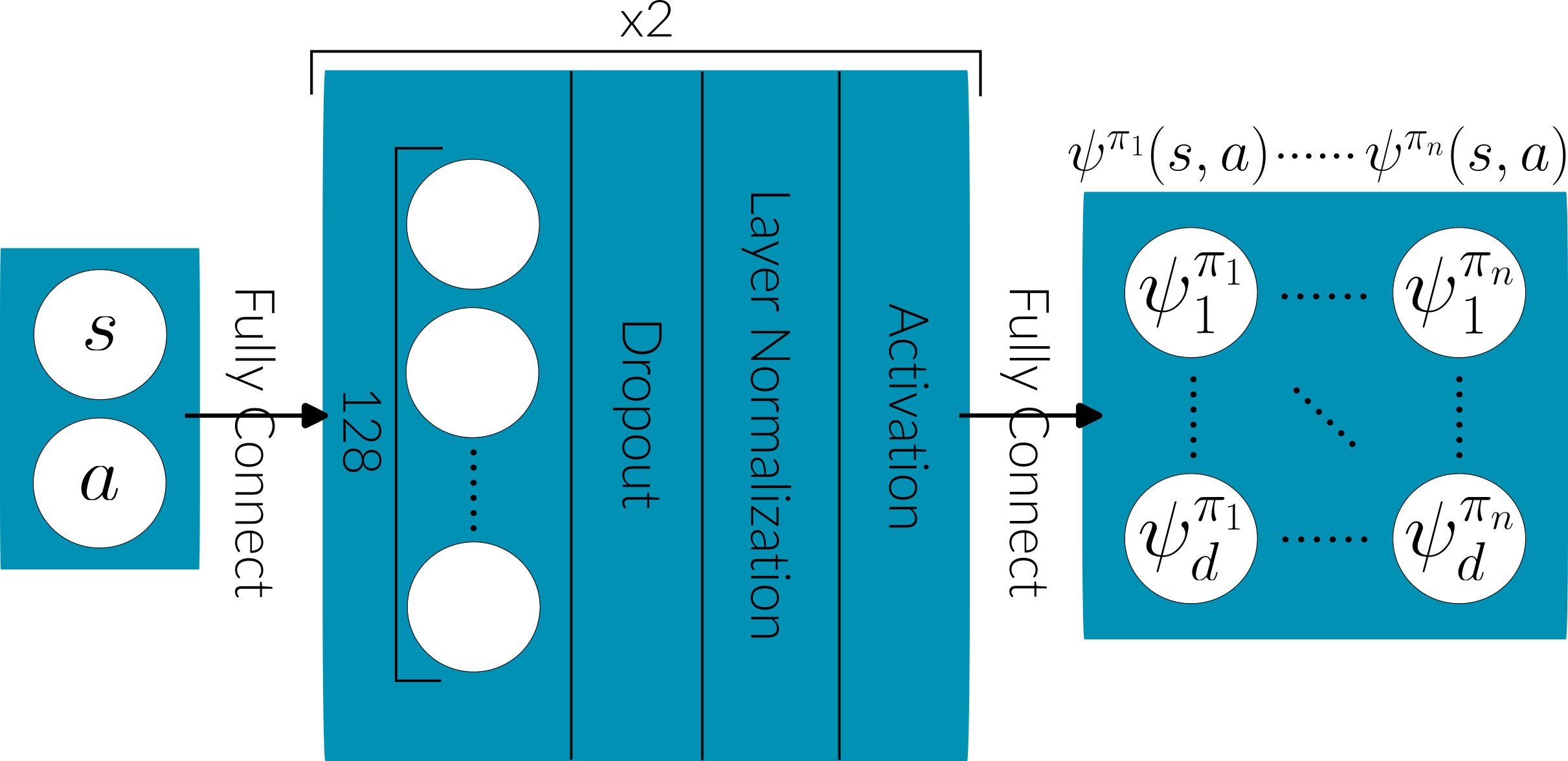}
		\caption{successor feature network.}
		\label{fig:sf_connect}
	\end{subfigure}
	\caption{network architecture. } 
	\label{fig:net_architecture}
\end{figure}

\begin{figure*}
	\centering
	\begin{subfigure}[t]{0.31\textwidth}
		\centering
		\includegraphics[width=0.95\textwidth]{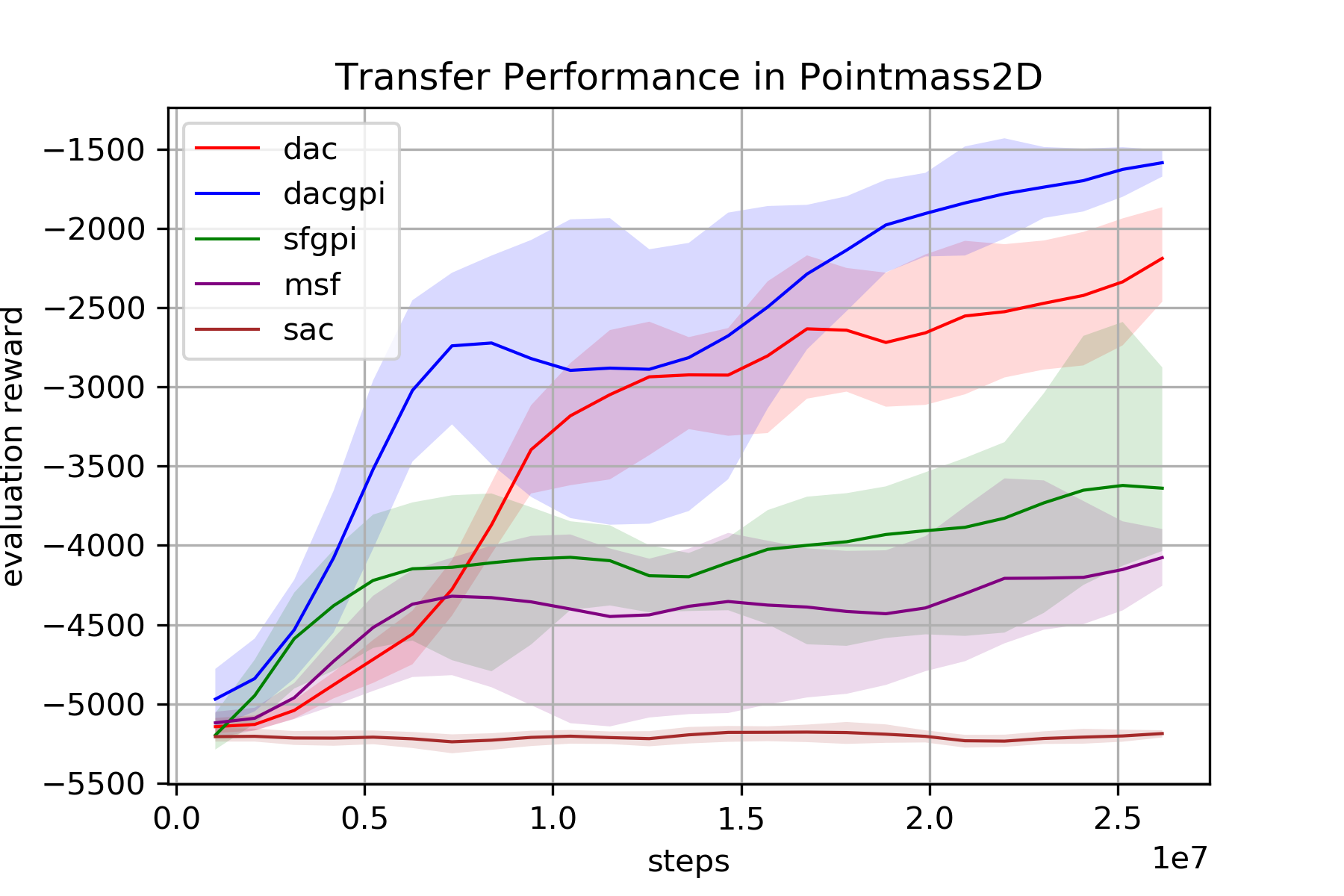}
		\caption{\textbf{Transfer performance}: Train with a velocity control task and evaluate with a mixture task $w^{mix}$ (~\ref{eqn:pm_weight}).}
		\label{fig:transfer_performance_pm2d}
	\end{subfigure}
	\hfill
	\begin{subfigure}[t]{0.31\textwidth}
	\centering
	\includegraphics[width=0.95\textwidth]{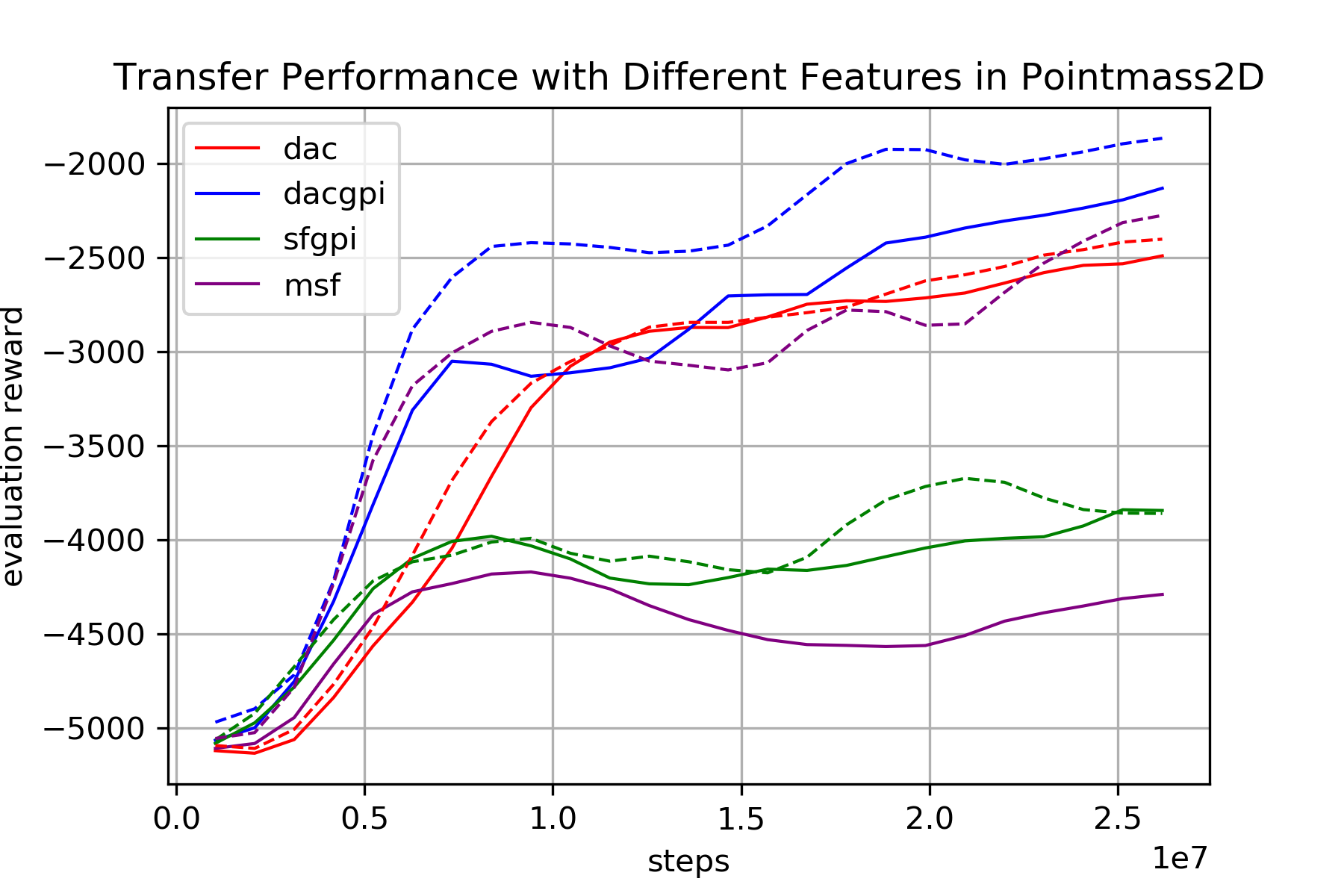}
	\caption{\textbf{Effect of feature sets in transfer}: Compared to regular features (solid), using augmented features (dotted) is beneficial.} 
	\label{fig:feature_set_pm2d}
	\end{subfigure}
	\hfill
	\begin{subfigure}[t]{0.31\textwidth}
		\centering
		\includegraphics[width=0.95\textwidth]{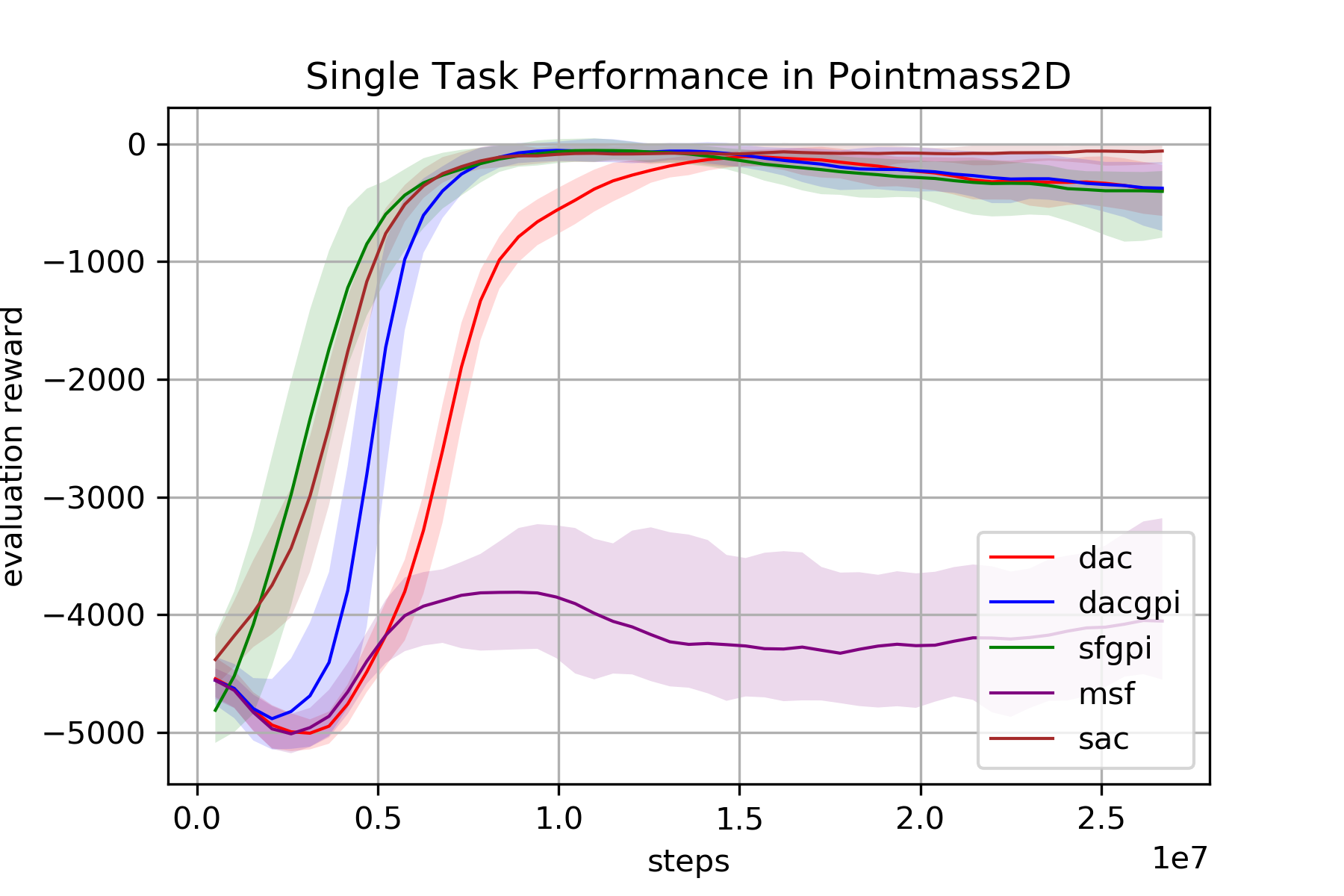}
		\caption{\textbf{Single task performance}: Due to the composition loss, composition agents have lower performance compared to the SAC in a single task $w=(\mathbf{1}, 0, 1)$.} 
		\label{fig:single_task_pm2d}
	\end{subfigure}
	\hfill
	\begin{subfigure}[t]{0.31\textwidth}
		\centering
		\includegraphics[width=0.95\textwidth]{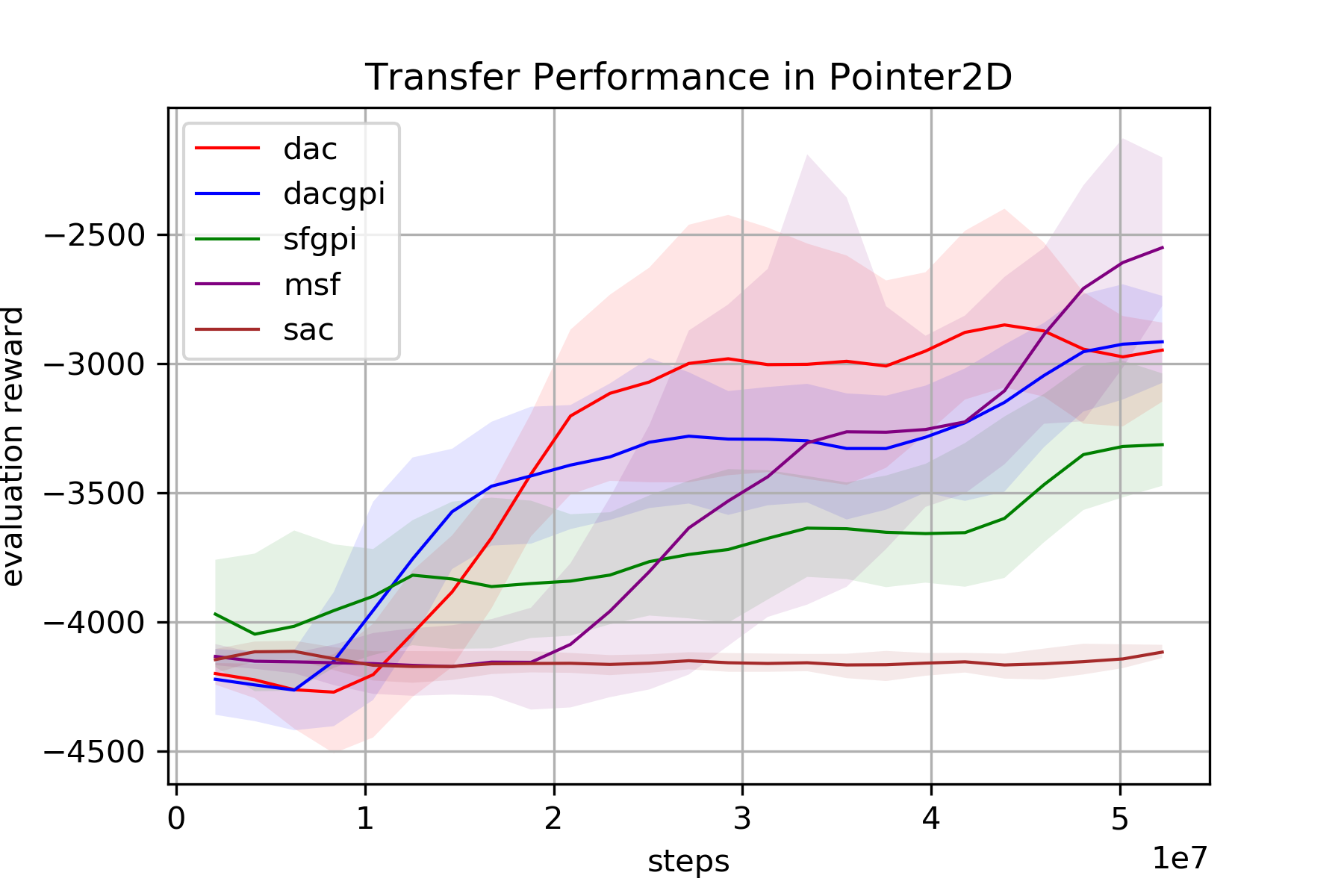}
		\caption{\textbf{Transfer performance}: Train with a velocity control task and evaluate with a mixture task $w^{mix}$ (~\ref{eqn:pm_weight}). }
		\label{fig:transfer_performance_pt2d}
	\end{subfigure}
	\hfill
	\begin{subfigure}[t]{0.31\textwidth}
		\centering
		\includegraphics[width=0.95\textwidth]{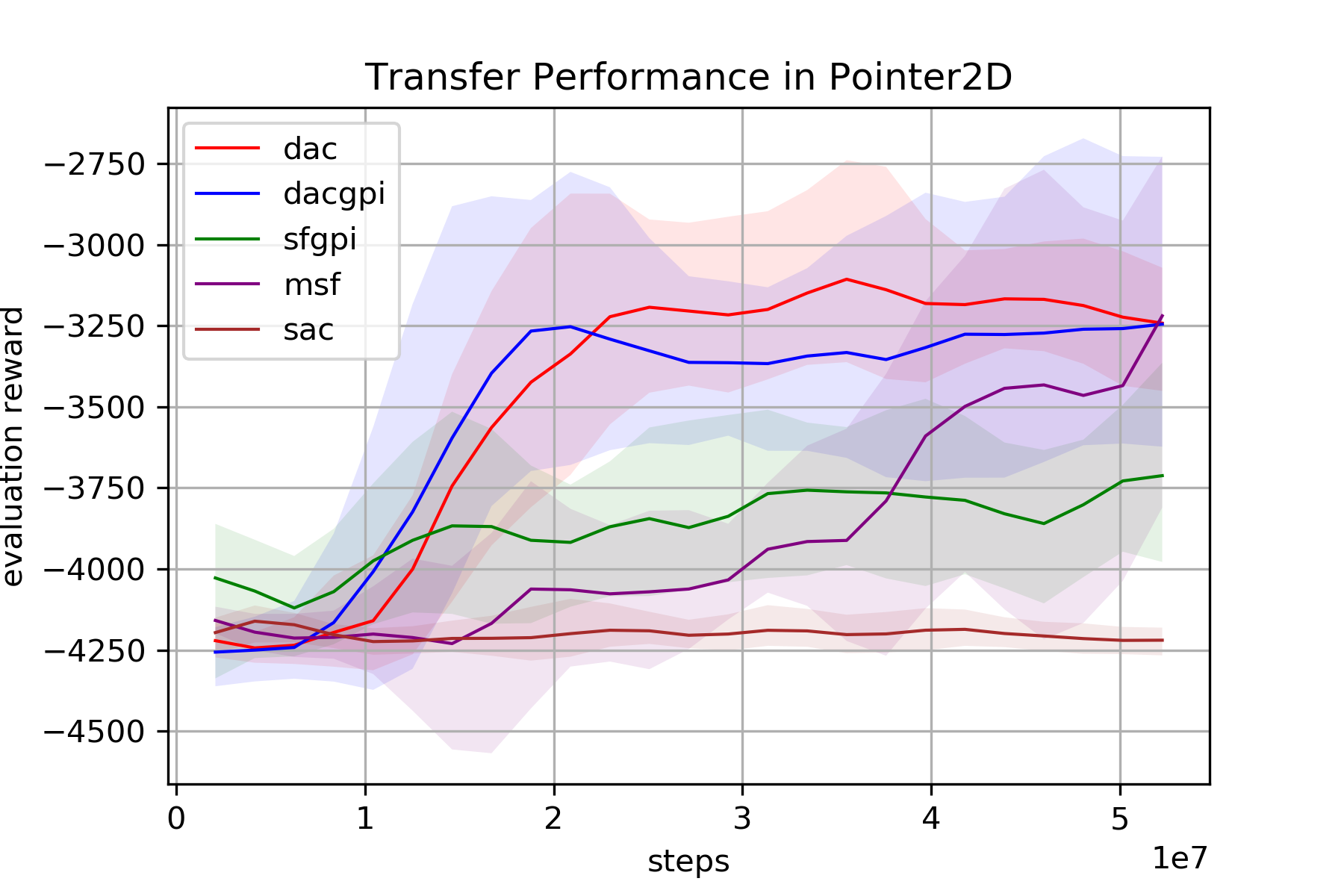}
		\caption{\textbf{Transfer performance}: Train with a velocity control task but evaluate with a position control task $w=(1, 0, 0, 0, 10)$.}
	\label{fig:transfer_task2_pt2d}
	\end{subfigure}
	\hfill	
	\begin{subfigure}[t]{0.31\textwidth}
		\centering
		\includegraphics[width=0.95\textwidth]{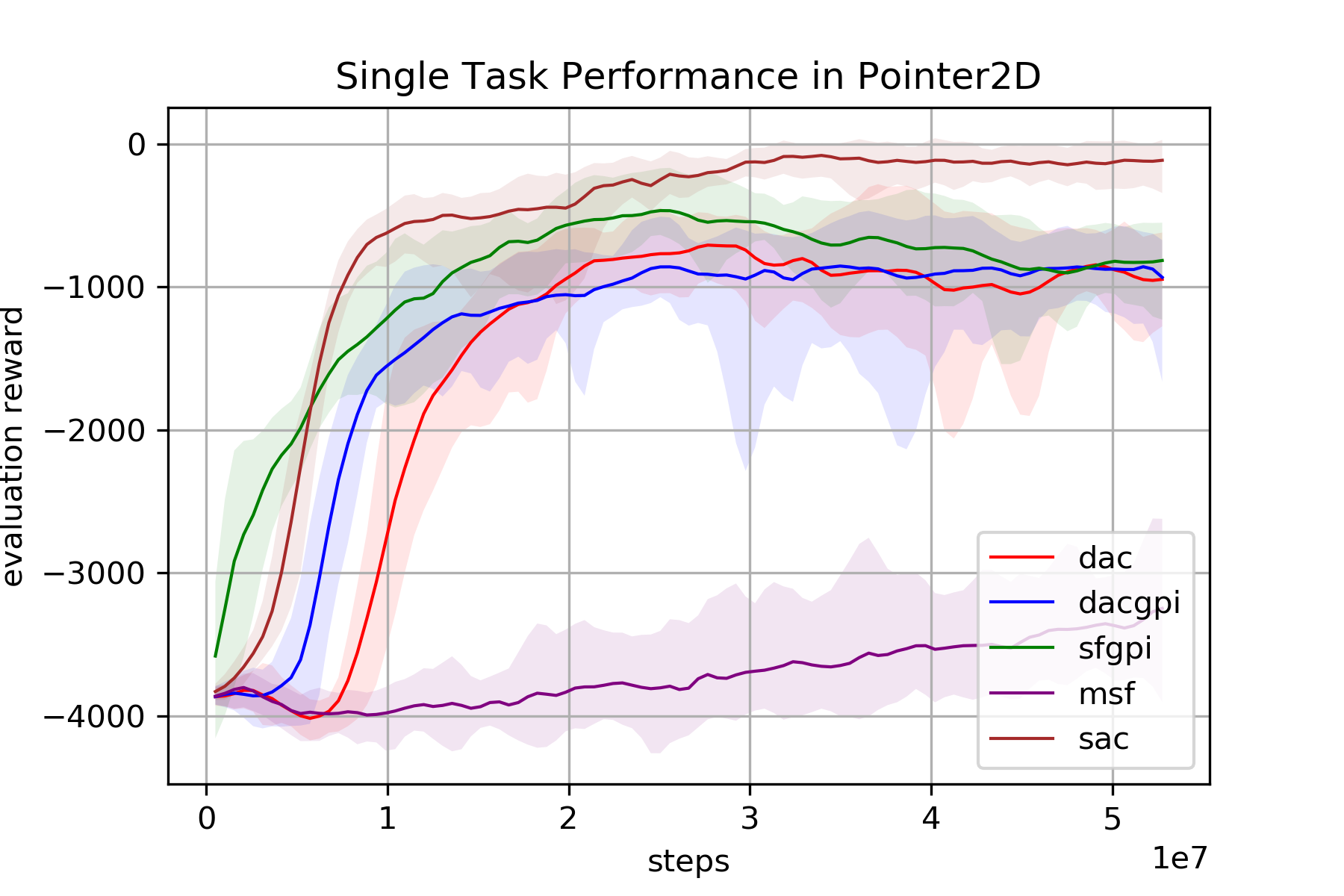}
		\caption{\textbf{Single task performance}: Due to the composition loss, composition agents have lower performance compared to the SAC in a single task $w=(1,0,0,0,1)$.}
		\label{fig:single_task_pt2d}
	\end{subfigure}
	\hfill
	
	\caption{Empirical results on Pointmass (top row) and Pointer (bottom row) show that the proposed composition agents can gradually improve transfer performance to unseen tasks and have comparable single-task performance to baseline SAC \cite{haarnoja2018soft}. SAC is presented in transfer tasks to show that tasks are not generalizable. Each curve represents the mean with two standard deviations of 5 experiments run by the best model from the hyper-parameter tuning. The tuning process is conducted by Bayesian optimization with hundred searches for each agent. Our ablation study indicates that dropout can hurt the performance while other techniques' effects remain unclear, including layer norm, prioritized experience replay, entropy tuning, and activation function.}
\end{figure*}

\subsection{Network Architecture}
We represent primitives and successor features by two neural networks as displayed in Fig.~\ref{fig:net_architecture}. Knowledge transfer can happen within the network since the layers or representations are shared. To reduce over-estimation in the SFs and enable higher update-to-data ratio (UTD ratio) \cite{hiraoka2021dropout}, we include \textit{dropout} and \textit{layernorm} to the SF network. To further mitigate the positive bias \cite{fujimoto2018addressing}, we apply the minimum double critic technique for the SF estimation, i.e., $\hat{\psi}^{\pi} = min(\psi_{\theta_1}^{\pi}, \psi_{\theta_2}^{\pi})$.

\subsection{Environment and Tasks}
Our training environment is implemented using IsaacGym, leveraging its extensive parallelization capabilities. While IsaacGym offers a range of pre-existing robots, we introduce abstract omni- and uni-directional robots, namely Pointmass and Pointer, designed to tackle our problem and gain valuable insights.

\subsubsection{State, Action, and Feature Space} 
The state and action space of the environments are presented in Table.~\ref{tab:env_space]}, 

\begin{table}[h]
	\centering
	\resizebox{0.4\textwidth}{!}{%
		\begin{tabular}{ c||c|c }
			\hline
			 &  \textit{Pointmass} & \textit{Pointer} \\
			\hline
			$\mathcal{A}$    & $(a_x, a_y, a_z)$ & $(a_{thrust}, a_{roll}, a_{pitch}, a_{yaw})$  \\
			\hline
			$\mathcal{S}$ & \multicolumn{2}{c}{$\Delta{\mathbb{X}}, \Delta{\mathbb{V}}, \Delta{\Theta}, \Delta{\Omega} \in \mathbb{R}^3$}   \\
			\hline
		\end{tabular}}
		\caption{ The action space for Pointmass are applied force on the x-, y-, and z-axis while for Pointer are applied thrust and attitude controls. Both environments share the same state space, namely relative position, velocity, orientation, and angular velocity between the robot and goal.
		\label{tab:env_space]} 
	}
\end{table}
Features, or sub-tasks, can be defined arbitrarily as long as they can be derived from the state space. We define three sets of features for the Pointmass and one for the Pointer, as shown in Table.~\ref{tab:feature_set}. The feature sets used in our study include the following variations: \textit{Regular}, which represents a standard approach for defining reward functions in robotics; \textit{Simple}, a simplified version used for demonstration purposes; and \textit{Augment}, which includes redundant features that generate additional primitives. For example, both $-|\Delta{\mathbb{X}}|$ and $-||\Delta{\mathbb{X}}||_2$ can achieve position control. Each primitive aims to maximize its corresponding feature or sub-task. In general, the task can be solved only when a sufficient amount of features are provided.
\begin{table}[h]
	\centering
	\resizebox{0.48\textwidth}{!}{%
		\begin{tabular}{ c||c|c }
			\hline
			\multirow{3}{*}{\textit{Pointmass}} & Regular & $-|\Delta{\mathbb{X}}|, -||\Delta{\mathbb{V}}||_2,  
			Success$  \\
			\cline{2-3}
			&Simple  & $-|\Delta{\mathbb{X}}|$  \\
			\cline{2-3}
			&Augment & $-|\Delta{\mathbb{X}}|, -||\Delta{\mathbb{X}}||_2, -|\Delta{\mathbb{V}}|, -||\Delta{\mathbb{V}}||_2,  
			Success$  \\
			\hline
			\textit{Pointer} & Regular & $-||\Delta{\mathbb{X}}||_2, -||\Delta{\mathbb{V}}||_2, 
			-||\Delta{\Theta}||_2,
			-||\Delta{\Omega}||_2,
			 Success$  \\
			\hline
	\end{tabular}}
	
	\caption{Environment feature sets. The absolute value and L2 norm are denoted as $|\cdot|$, $||\cdot||_2$, respectively. $Success$ is a binary feature triggered when the relative distance is less than 1 meter, i.e., $Success=1$ if $||\mathbb{X}||_2 <1$ else $0$. The default will be \textit{Regular} if not explicitly mentioned.  
	\label{tab:feature_set} }
\end{table}
\begin{figure}[h]
	\centering
	\begin{subfigure}[t]{0.235\textwidth}
		\includegraphics[width=1.0\textwidth]{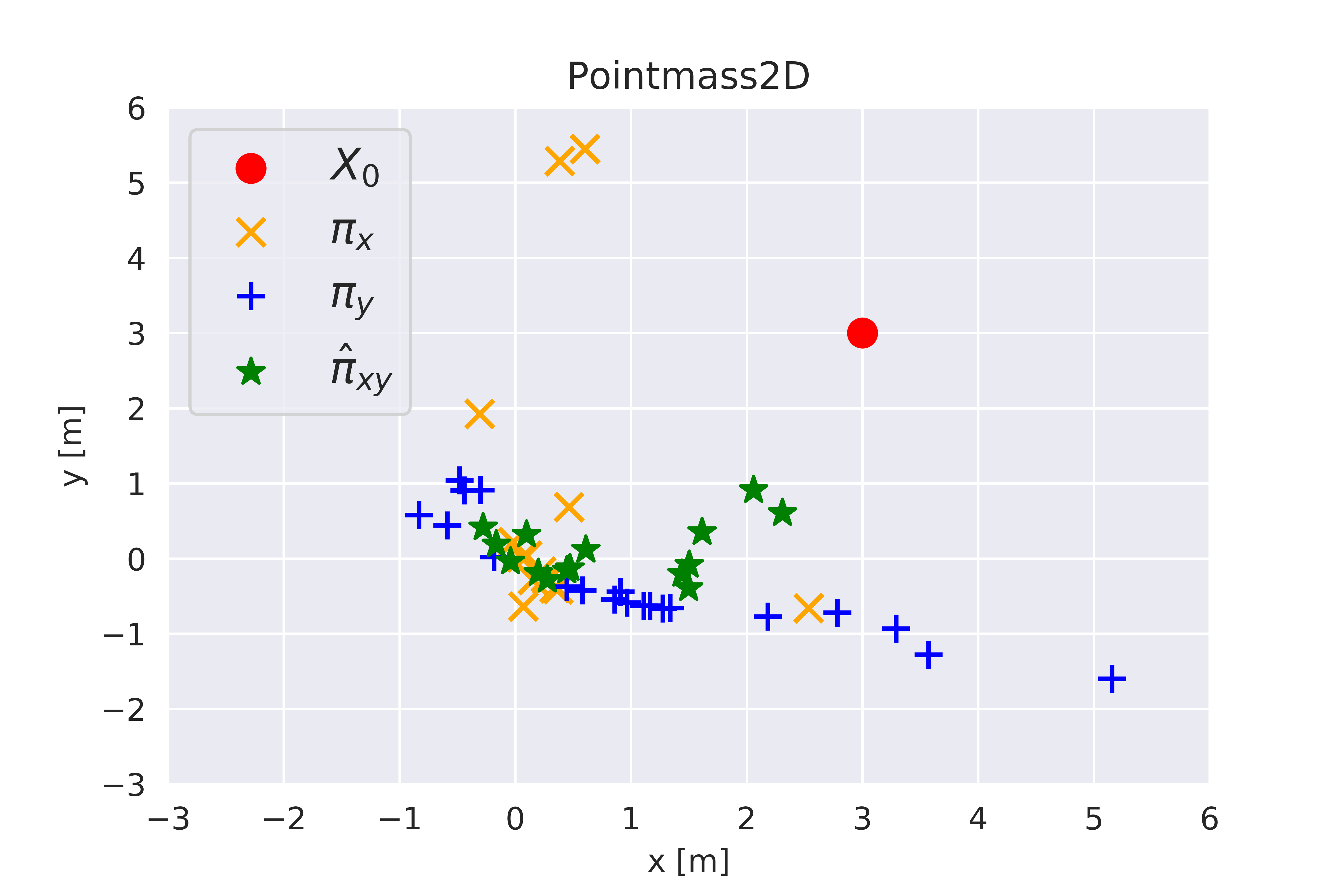}
		\caption{Sample the end position of 20 trajectories for x- (blue), y- (yellow) primitives and their composite (green). }
		\label{fig:primitives_pm2d}
	\end{subfigure}
	~
	\begin{subfigure}[t]{0.235\textwidth}
		\centering
		\includegraphics[width=1.0\textwidth]{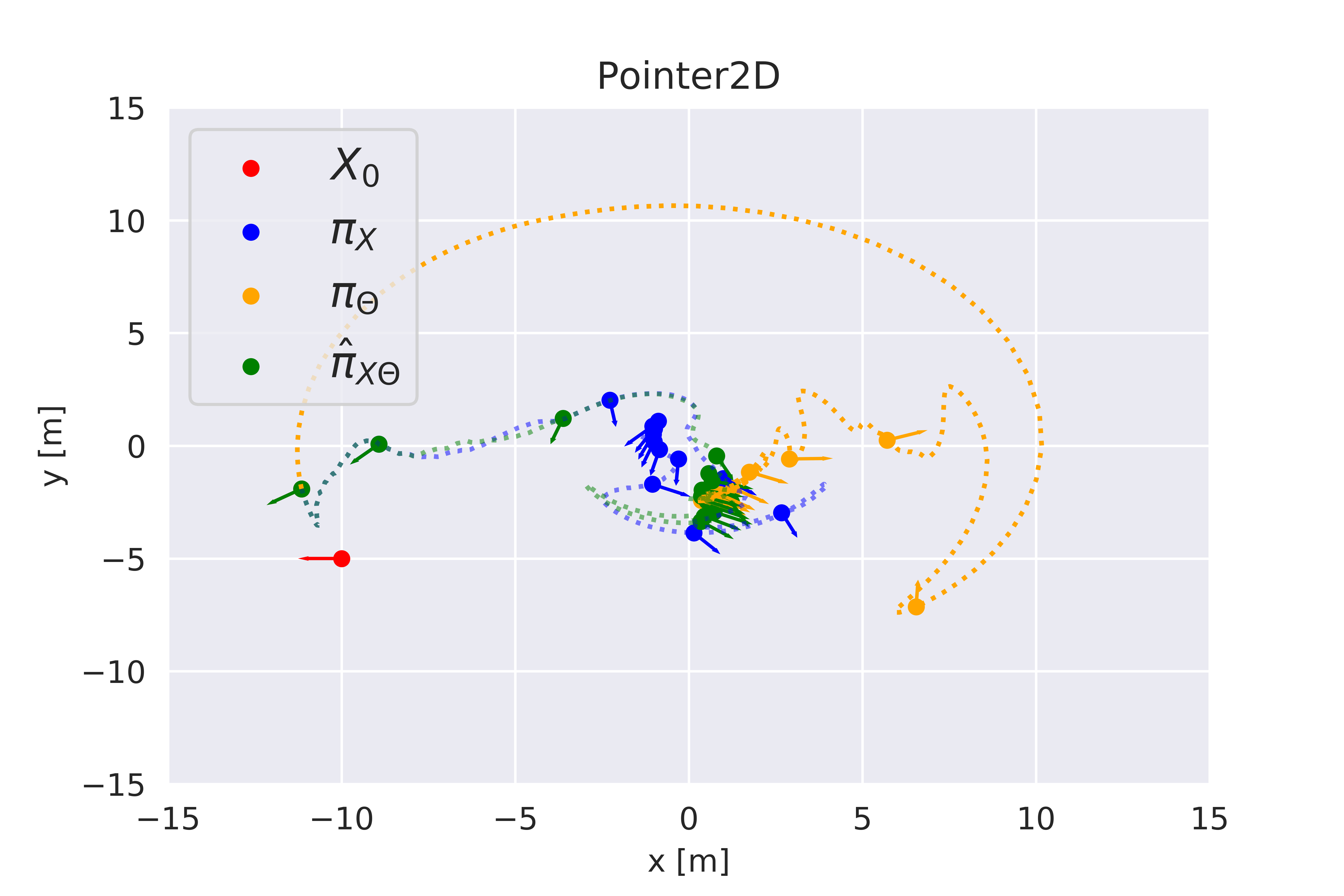}
		\caption{Sample one trajectory for the position (blue), angle (yellow) primitives, and their composite (green).}
		\label{fig:primitives_pt2d}
	\end{subfigure}
	\caption{Visualize primitive and composite distributions by sample trajectories starting from the initial position $X_0$ (red) to the goal at the origin. Composition agents allow solving new tasks by composing existing primitives. }
	\label{fig:primitive_composite_distribution}
\end{figure}

\subsubsection{Mixture Task}
\label{sec: Mixture Task}
We introduce a mixture task, which includes two tasks: \textit{navigation} and \textit{hover} tasks. The navigation task aims to bring the robot closer to the goal position, while the hover task requires the robot's velocity to approach the target velocity when near the goal. The task switches from navigation to hover when the distance to the goal is less than three meters. We can formulate this accordingly,  
\begin{equation}
	\begin{array}{l}
		w^{mix}=\begin{cases}
			w^{nav} & \text{if \ensuremath{d(s_{robot},s_{target})\geq 3m} }\\
			w^{hov} & \text{otherwise}.
		\end{cases}
	\end{array}\label{eqn:pm_weight}
\end{equation}
where by default, in Pointmass, $w^{mix}$ is set as $[(\mathbf{1}, 0, 10), (\mathbf{0}, 1, 10)]$ (bold symbols are vectors of the same length to the corresponding features). In Pointer, the default $w^{mix}$ is $[(1, .5, 0, 0, 10), (0, 1, 0, 0, 10)]$. The navigation task is essentially a position control task. In contrast, in hover tasks, the robots must maintain a circular motion to stay within the hover range and track the desired velocity simultaneously.

We assess transferability by employing different tasks for the training and evaluation phases. Specifically, we train primitives in a velocity control task, with $w_{train}=(\mathbf{0}, 1, 0)$ for Pointmass and $w_{train}=(0, 1, .5, 0, 0)$ for Pointer. Then, we evaluate the agents on the mixture task (Fig.~\ref{fig:transfer_performance_pm2d}~\ref{fig:transfer_performance_pt2d}) or a position control task (Fig.\ref{fig:transfer_task2_pt2d}). We observe that multi-task agents demonstrate greater resilience when the evaluation task differs from the training task. The impact of feature sets is illustrated in Fig.~\ref{fig:feature_set_pm2d}, where we find that using the Regular feature set is sufficient to solve the task, and additional features can further enhance transfer performance for MSF and DAC-GPI. Finally, Fig.~\ref{fig:primitive_composite_distribution} demonstrates the effectiveness of primitive composition in solving composite tasks that individual primitives alone cannot handle. We conclude that compositional agents transfer to different tasks if primitives are adequately trained. However, there is no performance guarantee for single task or transfer task due to the composition loss.

\begin{table}[h]
	\centering
	\resizebox{0.35\textwidth}{!}{%
		\begin{tabular}{ c||c }
			\hline
			\textit{composite} & \textit{action components $(a_x,a_y)$} \\
			\hline
			$\hat{\pi}^{MSF}$ & $  ( \mathcal{N}^{\pi_x}_x\epsilon, \mathcal{N}^{\pi_y}_y\epsilon)$   \\
			\hline
			$\hat{\pi}^{GPI}$ & $( \mathcal{N}^{\pi_x}_x\epsilon, \mathcal{N}^{\pi_y}_y)$ or $( \mathcal{N}^{\pi_x}_x, \mathcal{N}^{\pi_y}_y\epsilon)$   \\
			\hline
			$\hat{\pi}^{DAC}$ & $ ((\mathcal{N}^{\pi_x}_x)^{\kappa^{\pi_x}_x}(\epsilon)^{\kappa^{\pi_y}_x}, (\mathcal{N}^{\pi_y}_y)^{\kappa^{\pi_x}_y}(\epsilon)^{\kappa^{\pi_y}_y})$   \\
			\hline
	\end{tabular}}
	\caption{Composite distributions in Pointmass2D-Simple.  $\epsilon$ is the composition noise of unknown distribution. 
		\label{tab:composition noise} 
	}
\end{table}
\begin{figure}[h]
	\centering
	\begin{subfigure}[t]{0.23\textwidth}
		\includegraphics[width=1.0\textwidth]{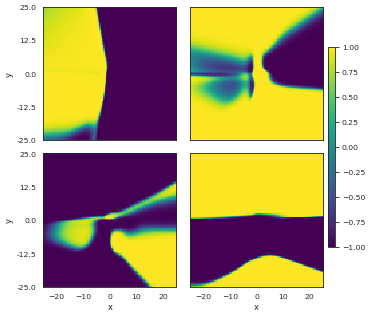}
		\caption{Top row: action components in $\pi_x=(a_x,a_y)$. Bottom row: $\pi_y$. Color bar range in $[-1,1]$. }
		\label{fig:primitives}
	\end{subfigure}
	~
	\begin{subfigure}[t]{0.23\textwidth}
		\centering
		\includegraphics[width=1.0\textwidth]{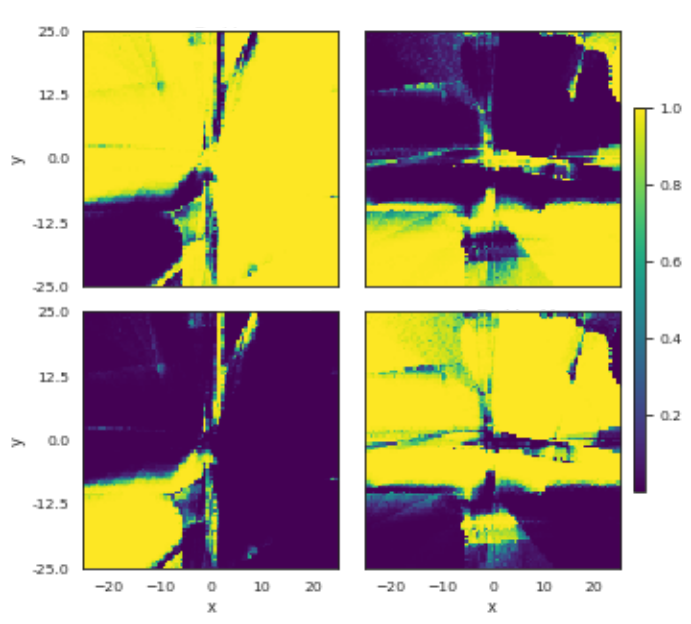}
		\caption{Top row: kappa $\kappa^{\pi_x}=(\kappa^{\pi_x}_x,\kappa^{\pi_x}_y)$. Bottom row: $\kappa^{\pi_y}$. Color bar range in $[0,1]$.}
		\label{fig:kappa}
	\end{subfigure}
	\caption{In Pointmass2D-Simple, DAC effectively removes compositional noise (Fig.(a) top right and bottom left) by reducing the corresponding $\kappa$. Each sub-figure represents a 2D plane within $[-20, 20]$ meters.}
	\label{fig:pm2_visual}
\end{figure}

\subsection{Composition Noise and Training Stability in Pointmass2D}

The presence of composition noise resulting from redundant action components in the primitive task reduces the training stability. In the Pointmass2D-Simple environment, the primitive actions corresponding to the features $-|\Delta{X}|$ and $-|\Delta{Y}|$ generate the distributions $\pi_x = ( \mathcal{N}^{\pi_x}_x, \epsilon)$ and $\pi_y= ( \epsilon, \mathcal{N}^{\pi_y}_y)$, introducing composition noise $\epsilon$ (Fig.~\ref{fig:primitives}) and composites (Table.~\ref{tab:composition noise}). In the case of MSF, the composite action contains noise in both components, while SF-GPI only contains noise in one component. Notably, DAC(-GPI) effectively mitigate noise in both action components, as the successor features reduce the impact of the noisy term (Fig.~\ref{fig:kappa}). The importance of filtering out composition noise for effective transfer is evident in Fig.~\ref{fig:transfer_performance_pm2d}. DAC(-GPI) exhibits the best asymptotic performance, followed by SF-GPI and MSF. We observed that composition noise hinders the MSF training stability (Fig.~\ref{fig:single_task_pm2d},~\ref{fig:single_task_pt2d}).

We conclude that applying MSF is impractical. Unlike other methods, which can handle any number of primitives, MSF requires that the feature dimension matches the number of primitives (i.e., n=d). Moreover, its optimistic assumption of linearity in both task and policy spaces is rarely valid.

\subsection{Single Task Performance}
Compared to baseline SAC \cite{haarnoja2018soft}, the multi-task agents exhibit lower asymptotic performance in the single task due to the composition loss (Fig.~\ref{fig:single_task_pm2d},~\ref{fig:single_task_pt2d}). We found that SF-GPI tends to explore less, potentially leading to faster yet premature convergence. We observed that the impact matrix estimation is often noisy, limiting DAC(-GPI)'s effectiveness. Hence, it may be beneficial to initially train primitives using alternative composition methods, transitioning to DAC(-GPI) once SFs achieve satisfactory precision and smoothness.


\section{Conclusions and Future Work}
\label{sec:5_discussion}
Our work unifies the SF-GPI and value composition to the continuous concurrent composition framework and allows the reconstruction of true policy from a set of primitives. The proposed method was extended to composition in the action space. In the Pointmass and Pointer environments, we demonstrate that our multi-task agents can approximate the true policy in real time and transfer the skills to solve unseen tasks while the single-task performance is competitive with SAC.
Our framework incorporates well with the reward-shaping techniques \cite{ng1999policy} and provides fine control over each sub-tasks via primitives. 
In addition, the task-agnostic property should benefit the autotelic agent \cite{colas2022autotelic} who can set goals and curriculum for themselves \cite{narvekar2020curriculum}. 
	
	\bibliographystyle{IEEEtran}
	\bibliography{biblio}
\end{document}